\documentclass{isprs}
\usepackage[margin=1in]{geometry} 
\usepackage{graphicx}             
\usepackage[colorlinks=false, pdfborder={0 0 0}]{hyperref} 
\usepackage{fontawesome} 
\usepackage{booktabs}
\usepackage{float}               
\usepackage{array}  
\usepackage{subcaption}
\usepackage{cite}
\usepackage[utf8]{inputenc}   
\usepackage[T1]{fontenc}      
\usepackage[margin=1in]{geometry}
\usepackage{graphicx}                
\usepackage[section]{placeins}       
\usepackage{caption}
\usepackage{comment}
\usepackage{soul} 
\usepackage{color} 
\usepackage{amsmath} 
\usepackage{hyperref}

\usepackage{fancyhdr}

\begin{document}





   



\title{50 Years of Water Body Monitoring:\\ The Case of Qaraaoun Reservoir, Lebanon}
\date{}



\author{
Ali Ahmad Faour\textsuperscript{1}, 
Nabil Amacha\textsuperscript{2}, 
Ali J. Ghandour\textsuperscript{3*}
}

\address{
\textsuperscript{1} American University of Beirut, Beirut, Lebanon \\
\textsuperscript{2} Lebanese University, Beirut, Lebanon \\
\textsuperscript{3} National Center for Remote Sensing, CNRS-L, Beirut, Lebanon \\
\textsuperscript{*} Corresponding Author: aghandour@cnrs.edu.lb

}

\icwg{}   

\abstract{
The sustainable management of the Qaraaoun Reservoir, the largest surface water body in Lebanon located in the Bekaa Plain, depends on reliable monitoring of its storage volume despite frequent sensor malfunctions and limited maintenance capacity. This study introduces a sensor-free approach that integrates open-source satellite imagery, advanced water-extent segmentation, and machine learning to estimate the reservoir’s surface area and, subsequently, its volume in near real time. Sentinel-2 and Landsat 1–9 images are processed, where surface water is delineated using a newly proposed water segmentation index. A machine learning model based on Support Vector Regression (SVR) is trained on a curated dataset that includes water surface area, water level, and water volume derived from a reservoir bathymetric survey. The model is then able to estimate the water body’s volume solely from the extracted water surface, without the need for any ground-based measurements. Water segmentation using the proposed index aligns with ground truth for over 95\% of the shoreline. Hyperparameter tuning with GridSearchCV yields an optimized SVR performance, with an error below 1.5\% of the full reservoir capacity and coefficients of determination exceeding 0.98. These results demonstrate the method’s robustness and cost-effectiveness, offering a practical solution for continuous, sensor-independent monitoring of reservoir storage. The proposed methodology is applicable to other water bodies and generates over five decades of time-series data, offering valuable insights into climate change and environmental dynamics, with an emphasis on capturing temporal trends rather than exact water volume measurements.
}


\maketitle

\section{Introduction}
The Qaraaoun Reservoir (QR) is located at an average altitude of approximately 850 meters in the central Bekaa Valley, between the Mount Lebanon range and the Anti-Lebanon range. The reservoir lies near the Qaraaoun village, specifically between the following geographic coordinates: 33°35'37"N, 33°32'53"N and 35°40'56"E, 35°42'26"E, and was formed in 1959 by the construction of the Qaraaoun Dam across the Litani River. With a storage capacity exceeding 220 million cubic meters, the reservoir receives water from snow, rainfall, and several existing springs, serving as a vital water resource for the region and supplying water to about one million people. This large volume buffers seasonal rainfall variability and provides a stable water supply that supports extensive irrigation systems, enhancing agricultural productivity in the valley. During extreme droughts, the reservoir plays a critical role in meeting domestic and industrial water demands. Additionally, its storage capacity contributes to hydroelectric power generation, providing approximately 8\% of Lebanon’s electricity, making it an essential component of regional economic development. These factors underscore the importance of consistently monitoring and managing the reservoir’s water volume to ensure long-term sustainability and resilience amid declining water availability and climate change~\cite{1}\cite{2}.

The Litani River Authority (LRA) is a governmental agency with key responsibilities related to the management of the Qaraaoun Reservoir (QR) and the broader Litani River system. Its mandate includes monitoring water quantity through a network of gauging stations installed on major Lebanese rivers and their tributaries. The LRA is also authorized to monitor surface water across the national territory and has recently initiated groundwater monitoring within the Litani Basin. In addition, the agency regulates water distribution to meet agricultural, industrial, and domestic needs, supplies irrigation water to farmers located south of the Damascus Road and in southern Lebanon, and contributes to hydropower production. Despite these extensive responsibilities, the LRA does not yet have full management authority over the entire Litani Basin.

Specifically, for the QR, the LRA employs an integrated network of hydrometric stations to continuously monitor the volume of water. However, challenges such as sensor malfunction and human error pose significant risks to accurate data collection, potentially compromising effective water management. In many developing countries, the maintenance of sensor-based systems often falls short of established standards and best practices due to financial constraints and a shortage of experienced personnel in government institutions. As a result, accurate recording of water volume is often compromised.


To address these challenges, this article proposes an efficient sensor-free approach that leverages remote sensing technology and machine learning to generate accurate weekly estimates of water volume in the Qaraaoun Reservoir. The contribution of this paper is threefold, as follows:

\begin{itemize}
  \item A machine learning model was developed to infer the reservoir's water volume in near real time from Sentinel-2 and Landsat 1–9 imagery. The model takes the extracted water surface area as input and estimates the corresponding water volume without relying on any ground-based sensor readings.
  \item Because the model’s performance is highly sensitive to the accuracy of water surface extraction, we introduce a novel water segmentation index that combines two existing indices from the literature using a weighted sum.
  \item Finally, we developed an interactive, web-based platform to visualize volume trends and segmentation results. The dashboard hosts a time series of over 50 years of reservoir statistics and serves as a valuable tool for researchers and stakeholders to explore environmental patterns and study the impacts of climate change. The dashboard is accessible at: \url{https://geoai.cnrs.edu.lb/qaraaoun}.
\end{itemize}

\begin{figure}[htbp]
    \centering
    \includegraphics[width=0.95\linewidth]{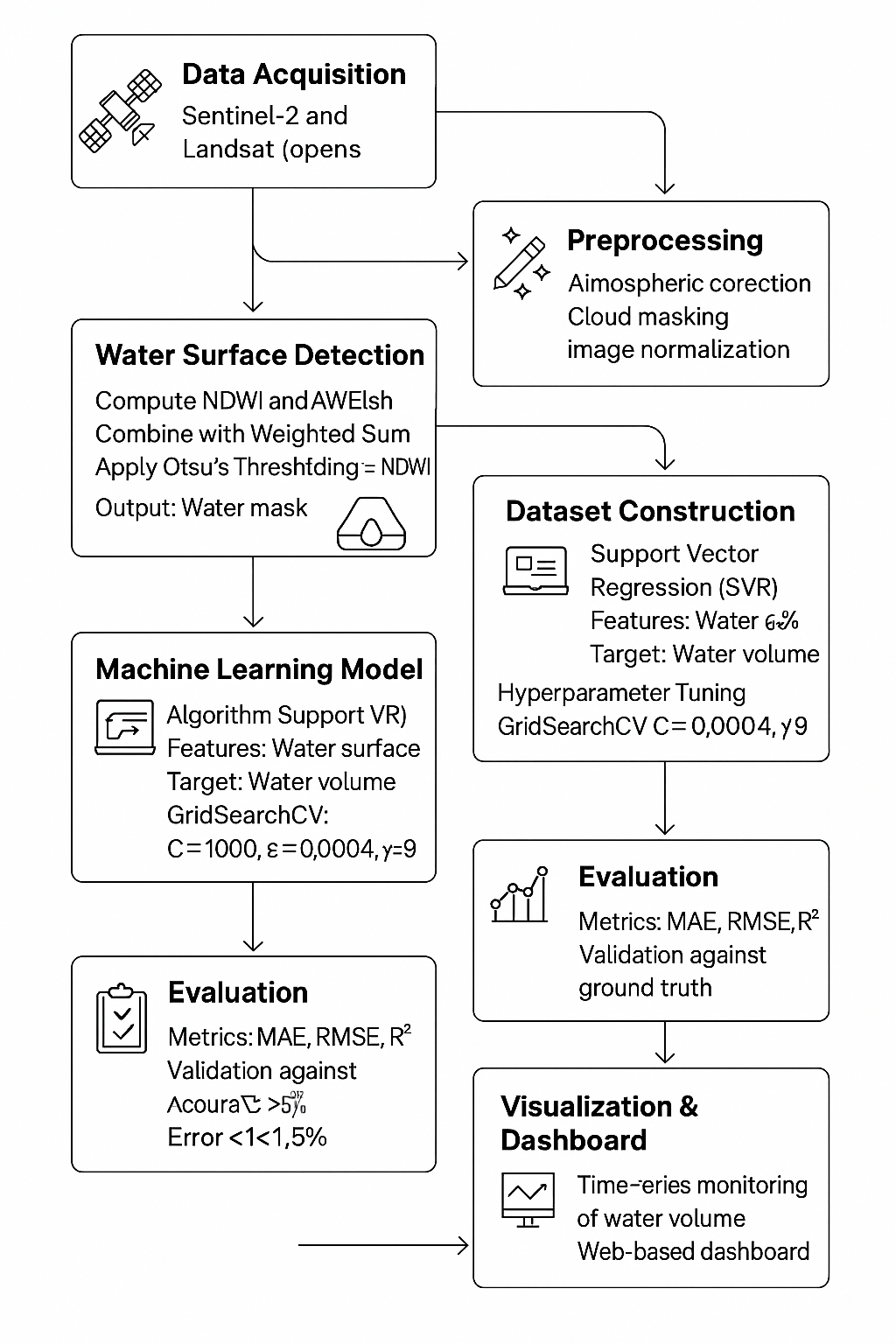}
    \caption{Workflow of the proposed pipeline, from data acquisition and preprocessing to water surface detection and volume estimation.}
    \label{fig:workflow}
\end{figure}

The proposed architecture is illustrated in Figure~\ref{fig:workflow}, outlining the complete workflow—from data acquisition and preprocessing to water surface detection, machine learning-based volume estimation, and dashboard visualization. Although applied here to QR, this architecture is adaptable and can be replicated for any other water body.

The rest of this paper is organized as follows: Section 2 discusses water spectral indices and introduces a very accurate water segmentation index. Section 3 outlines the implementation of the Support Vector Regression (SVR) model for volume estimation and the data preparation process using bathymetric surveys. Section 4 presents both qualitative and quantitative findings, emphasizing segmentation accuracy and the model's performance based on various metrics. The Conclusion in Section 5 underscores the effectiveness of the proposed sensor-free monitoring approach while acknowledging potential areas for improvement.


\section{Water Index}
\begin{figure*}[t]
    \centering
        \begin{subfigure}[t]{0.3\textwidth}
        \centering
        \includegraphics[width=\textwidth, height=10cm]{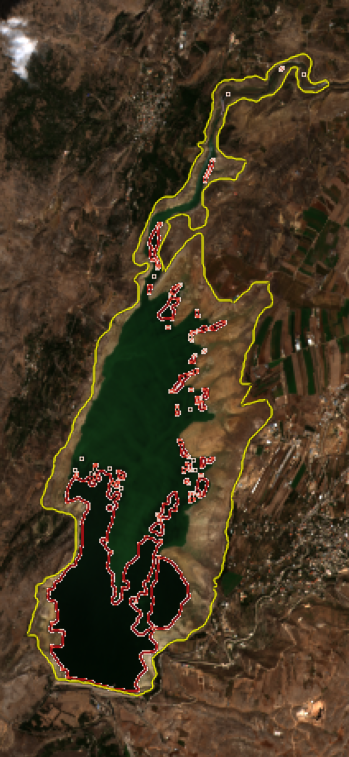}
        \caption{$NDWI$ water segmentation results.}
        \label{fig:ndwi}
    \end{subfigure}
    ~ 
    \begin{subfigure}[t]{0.3\textwidth}
        \centering
        \includegraphics[width=\textwidth, height=10cm]{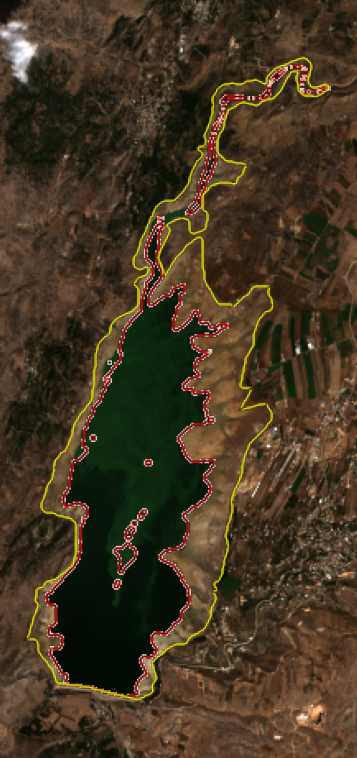}
        \caption{$AWEInsh$ water segmentation results.}
        \label{fig:aweinsh}
    \end{subfigure}%
    ~ 
    \begin{subfigure}[t]{0.3\textwidth}
        \centering
        \includegraphics[width=\textwidth, height=10cm]{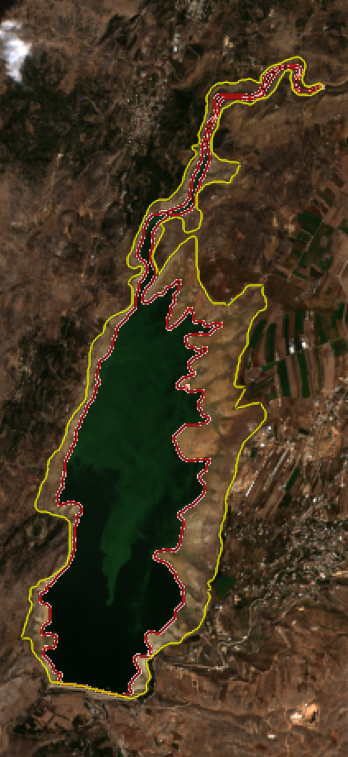}
        \caption{$WCWI$ water segmentation results.}
        \label{fig:composite}
    \end{subfigure}%
    \caption{
        Comparison of water segmentation results using different water indices: 
        (a) $NDWI$, (b) $AWEInsh$ and (c) $WCWI$ composite index on a Sentinel-2 imagery from 17 October 2023 where yellow outlines represent the nominal lake contour and red indicate detected water extent.
    }
    \label{fig:water_indices_comparison}
\end{figure*}

The detection of the water surface is performed using a set of spectral indices supplemented with Otsu’s adaptive thresholding. The most widely used index in the literature is the normalized difference in water index ($NDWI$), which is calculated as shown in Equation~\ref{eq:ndwi}:

\begin{equation}
\label{eq:ndwi}
NDWI = \frac{Green - NIR}{Green + NIR}
\end{equation}

where $NDWI$ varies from –1 to +1 and the existence of water will yield a positive value between zero and one. The primary use of $NDWI$ is to refine water pixel values to an extreme yielding a bimodal distribution; subsequently, because of the way new pixel values are distributed, Otsu's thresholding technique will be able to successfully capture an optimal threshold for separating water from non-water pixels.

Qaraaoun reservoir has an abundant presence of vegetation and soil, which can be mistaken for water. $NDWI$ is effective in distinguishing areas of vegetation and soil from water. Water typically exhibits low reflectance in both Green and NIR bands, leading to high values of $NDWI$, making it easier to distinguish it from other land covers.


While $NDWI$ is widely used in the literature, multiple studies have shown that spectral indices like $MNDWI$~\cite{9} and $ANDWI$~\cite{10} might outperform $NDWI$ in faithfully separating water pixels from noise especially in turbid and built-up areas. A careful choice of spectral indices is indeed pivotal for accurate water detection. 

The effectiveness of water indices for detecting surface water extent varies depending on seasonal and geographic factors. The Qaraaoun Reservoir experiences significant shadowing effects from surrounding mountain chains that alter the optical properties of satellite images. This leads to inaccurate classification, as pixels in shadowed regions exhibit low spectral reflectance. These non-water properties can reduce the accuracy of existing water indices, leading to segmentation errors at different times of the year. Therefore, selecting the most appropriate water index is crucial to ensuring reliable surface water extent detection.

In other words, non-water pixels and water pixels may have similar spectral reflectance. However, while shadowing can pose a problem, its impact is relatively minimal compared to that of soil and vegetation, which significantly hinders accurate segmentation. This observation makes the Automated Water Extraction Index – Non-Shadow ($AWEInsh$) a natural choice, as it is specifically designed to enhance the detection of water features in satellite images. It leverages multiple spectral bands, including the Near Infrared (NIR) and Shortwave Infrared (SWIR) bands, to maximize contrast between water and non-water features ~\cite{10}. The mathematical formulation for calculating $AWEInsh$ is shown in Equation~\ref{eq:aweinsh}:

\begin{multline}
\label{eq:aweinsh}
AWEInsh = 4 \times (Green - \text{SWIR}_1) - \\ (0.25 \times NIR + 2.75 \times \text{SWIR}_2)
\end{multline}


To make use of both $NDWI$ and $AWEInsh$ indices, a weighted sum referred to as Weighted Composite Water Index ($WCWI$) is proposed here.
The newly proposed water index is illustrated in Equation~\ref{eq:WCWI}:
\begin{equation}
\label{eq:WCWI}
WCWI = 0.8 \times AWEInsh + 0.2 \times NDWI
\end{equation}

Results in Figure~\ref{fig:water_indices_comparison} show the performance of the three indices $NDWI$, $AWEInsh$ and $WCWI$ on a Sentinel-2 for the QC dated on 17 October 2024. It is clear in Figure~\ref{fig:water_indices_comparison}(\subref{fig:ndwi}) that $NDWI$ struggles to segment water extent when color varies sharply in the reservoir. $AWEInsh$ on the other hand results in few segmentation errors towards the center of the reservoir. It also suffers from under-segmentation at the upper narrow upstream channel of the lake within shallow water and minimal width although not easily visible to the naked eye. The under-segmentation in Figure~\ref{fig:water_indices_comparison}(\subref{fig:aweinsh}) reduces the total water surface area, consequently leading to an underestimation of water volume. in contrast, the proposed $WCWI$ produces an accurate water segmentation mask, as demonstrated in Figure~\ref{fig:water_indices_comparison}(\subref{fig:composite}). This analysis was performed on dozens of images across the time-series; however, due to space limitations, we present only one representative example here and additional results are provided in the Results section.

\section{Model Design}
In an effort to improve water management in the Qaraaoun Reservoir, the Litani River Basin Management Support (LRBMS) program was initiated in 2013 and conducted a bathymetric survey of the reservoir. The survey aimed to assess sedimentation by comparing recent depth data with a topographic map from 1950. The survey was carried out through a series of east–west and west–east transects using a boat equipped with a Doppler flow meter (River Surveyor). This process produced high-resolution depth profiles used to establish an updated level–volume curve, serving as the definitive ground truth for all subsequent volume estimations~\cite{5}.

\begin{figure*}[t]
    \centering
    \begin{subfigure}[t]{0.21\textwidth}
        \centering
        \includegraphics[height=6cm]{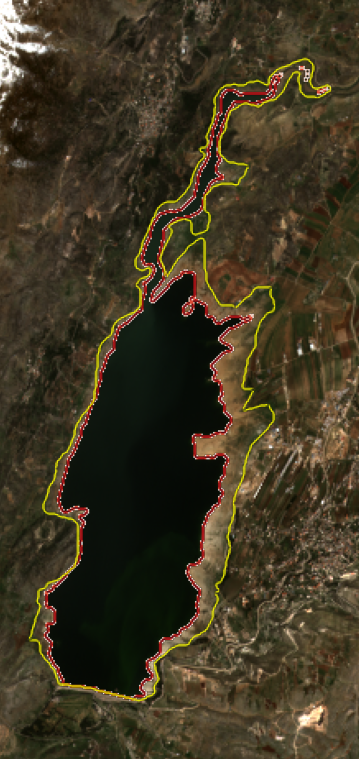}
        \caption{}
        \label{fig:image4a}
    \end{subfigure}
    ~ 
    \begin{subfigure}[t]{0.21\textwidth}
        \centering
        \includegraphics[height=6cm]{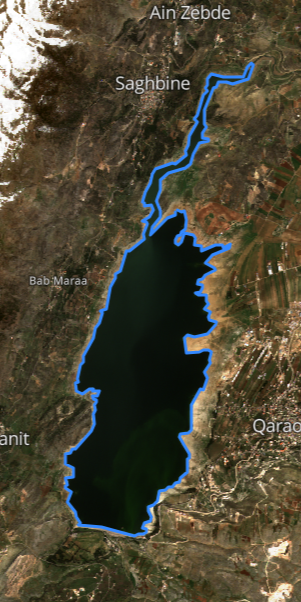}
        \caption{}
        \label{fig:image4b}
    \end{subfigure}%
    ~ 
    \begin{subfigure}[t]{0.21\textwidth}
        \centering
        \includegraphics[height=6cm]{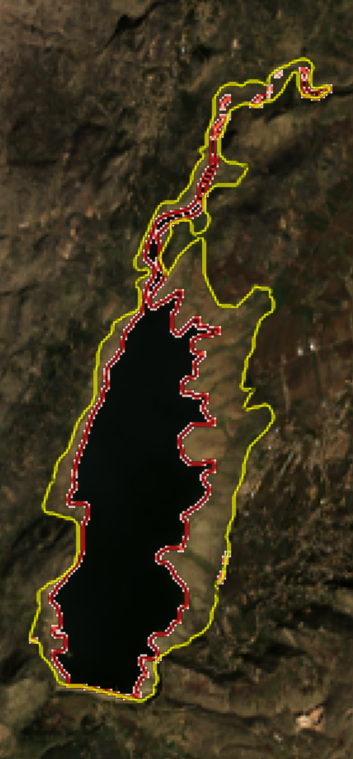}
        \caption{}
        \label{fig:image5a}
    \end{subfigure}%
    ~ 
    \begin{subfigure}[t]{0.21\textwidth}
        \centering
        \includegraphics[height=6cm]{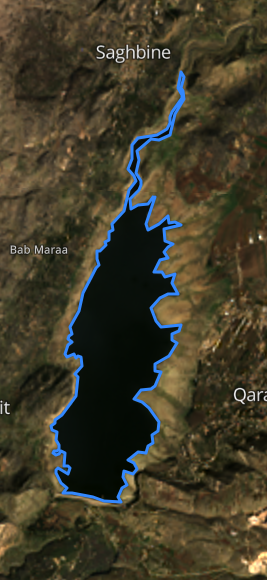}
        \caption{}
        \label{fig:image5b}
    \end{subfigure}%
    \\
    \begin{subfigure}[t]{0.21\textwidth}
        \centering
        \includegraphics[height=6cm]{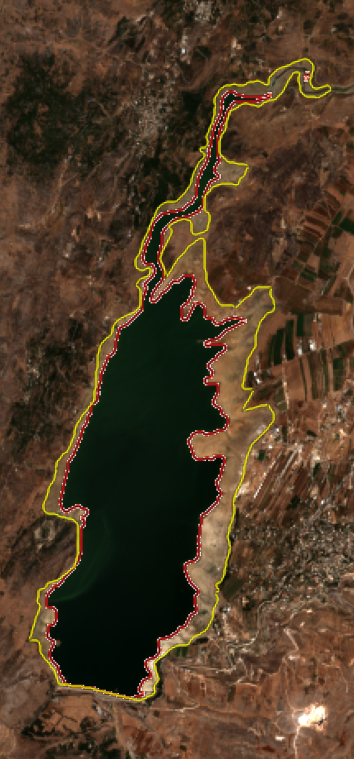}
        \caption{}
        \label{fig:image7a}
    \end{subfigure}%
    ~ 
    \begin{subfigure}[t]{0.21\textwidth}
        \centering
        \includegraphics[height=6cm]{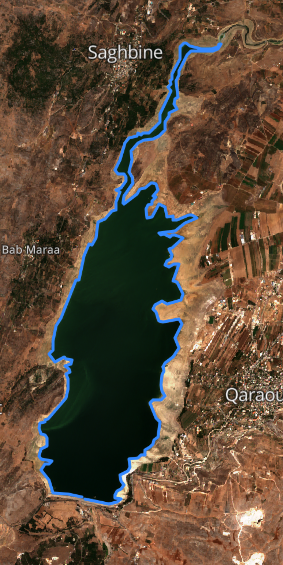}
        \caption{}
        \label{fig:image7b}
    \end{subfigure}%
    ~
    \begin{subfigure}[t]{0.21\textwidth}
        \centering
        \includegraphics[height=6cm]{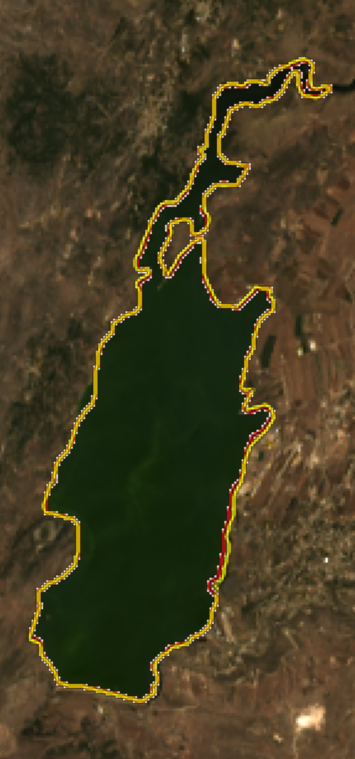}
        \caption{}
        \label{fig:image8a}
    \end{subfigure}%
    ~ 
    \begin{subfigure}[t]{0.21\textwidth}
        \centering
        \includegraphics[height=6cm]{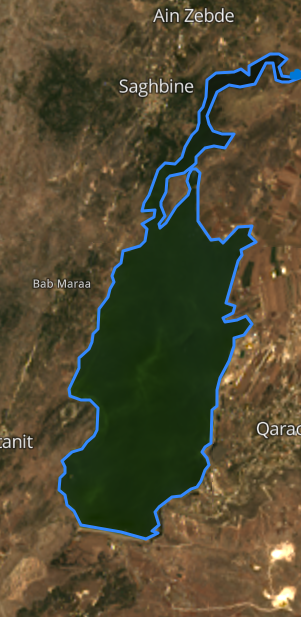}
        \caption{}
        \label{fig:image8b}
    \end{subfigure}%
    \caption{Water segmentation results in red versus ground truth values in blue: (a) and (b) Sentinel-2 imagery for 1 March 2023; (c) and (d) Landsat 8 imagery for 21 January 2023; (e) and (f) Sentinel-2 imagery for 26 September 2024; and (g) and (h) Landsat 8 imagery for 8 June 2024. The nominal lake contour is shown in yellow.}
    \label{fig:l8_eval}
\end{figure*}

Existing bathymetric data were first digitized, geo-referenced, and transformed into a digital elevation model (DEM). Discrete depth measurements were then interpolated using the “Nearest Neighbor” method to create a continuous surface representing the lakebed elevation. Using the DEM, a simulation of various water levels that the reservoir might experience was performed. 

We then utilized ground-truth data collected from the hydrometric station on the QR to construct a dataset comprising: \textit{(i)} water level measurements obtained from on-site sensors, \textit{(ii)} water surface area extracted from satellite imagery using the proposed water index, and \textit{(iii)} water volume estimates computed using the water level, surface area, and reservoir bathymetry. This dataset was then used to train a machine learning model capable of inferring water volume solely from water surface input, thereby eliminating the need for water-level measurements and enabling a sensor-free estimation approach.

Water bodies tend to fill in a nonlinear manner due to their complex geometry; thus, Support Vector Regression (SVR) was selected as the most suitable model for describing the correlation between water surface percentage and water volume. In supervised learning, the SVR model was trained on the constructed dataset, with the water surface percentage as the input feature and relative water volume as the output target. The data were divided into training and testing sets, where 80\% of the data were used for training and 20\% for testing. To enhance model convergence and accuracy, a Min–Max scaler was applied to both sets to normalize data values within the range [0,1]. An SVR model with an RBF kernel and standard hyperparameters was first instantiated.

Subsequently, a thorough hyperparameter optimization process was conducted using GridSearchCV with 10-fold cross-validation. This process tuned the key parameters relative to the mean squared error criterion and yielded optimal hyperparameter values of $C = 1000$, $\epsilon = 0.0004$, and $\gamma = 9$. The model with these settings achieved an average Mean Absolute Error ($MAE$) of approximately 0.0122 and a Root Mean Squared Error ($RMSE$) of approximately 0.0216 over the test set. The $MAE$ provides an average estimate of the absolute difference between the predicted and actual values, demonstrating high precision, while the $RMSE$, which penalizes larger errors more heavily, captures the model’s ability to fit the underlying nonlinear relationship. This process of training and tuning produced a robust inference model for accurate water volume estimation in the Qaraaoun Reservoir. Additional results are presented in the following section.

\section{Results}
\begin{figure*}[t]
    \centering
    \includegraphics[width=0.95\textwidth]{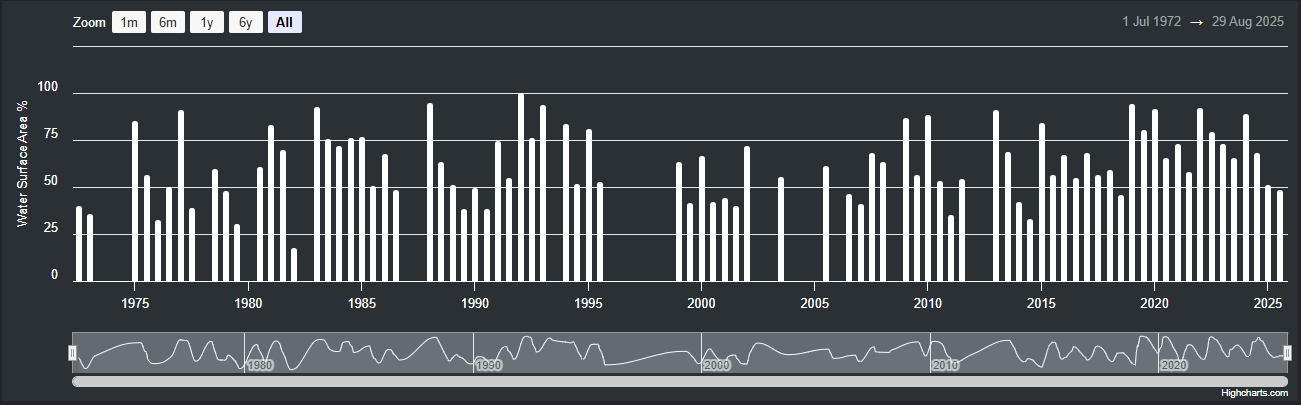}
    \caption{Time series (1973–2025) of water surface area and storage volume in the Qaraaoun Reservoir. July 2025 recorded an exceptionally low volume ($49.2\times10^6\,$m³), representing a 66\% decline from July 2024 and 56\% from July 2023, highlighting an emerging drought signal (see \href{https://geoai.cnrs.edu.lb/qaraaoun}{[Link]}).}

    \label{fig:50year_progression}
\end{figure*}

Results were gathered and analyzed to evaluate the segmentation and volume estimations obtained from the proposed solution presented in this paper. Segmentation accuracy was assessed by comparing the water-segmented imagery produced by the previously described algorithm with ground truth imagery. In Figure~\ref{fig:l8_eval}, Sentinel-2 and Landsat-8 imagery were used for qualitative analysis. Figures~\ref{fig:l8_eval}(\subref{fig:image4a}), \ref{fig:l8_eval}(\subref{fig:image5a}), \ref{fig:l8_eval}(\subref{fig:image7a}), and \ref{fig:l8_eval}(\subref{fig:image8a}) show that the proposed approach accurately captured the lake’s water extent with high precision (exceeding 95\% along the shoreline) when compared with the ground truth shown in Figures~\ref{fig:l8_eval}(\subref{fig:image4b}), \ref{fig:l8_eval}(\subref{fig:image5b}), \ref{fig:l8_eval}(\subref{fig:image7b}), and \ref{fig:l8_eval}(\subref{fig:image8b}), respectively. The yellow contour represents the lake’s nominal boundary, the red outline indicates the water segmentation result, and the blue contour depicts the manually labeled ground truth water mask. Small under-segmented gaps appear in the narrow upstream channel north of Saghbine and along the eastern shore. Shallow, turbid water and wet soil occasionally lead to slight under-segmentation, which was substantially minimized in this study by employing the proposed $WCWI$ index.

\begin{figure}[htbp]
    \centering
    \includegraphics[width=1\linewidth]{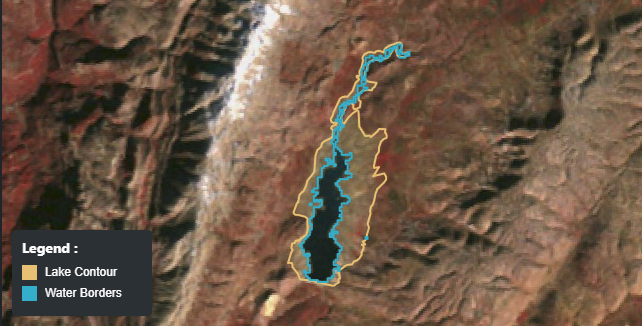}
    \caption{Historical Landsat-3 satellite image of the Qaraoun Reservoir captured on 17 January 1982, believed to represent the minimum recorded water extent in the past five decades.}
    \label{fig:min}
\end{figure}

The volume time series presented in Figure~\ref{fig:50year_progression} illustrates the percentage of water surface area and volume estimations over a period exceeding 50 years for the Qaraaoun Reservoir. The maximum recorded surface water coverage occurred on \textbf{23 May 1992}, with an estimated water volume of \textbf{182{,}802{,}172~m\textsuperscript{3}}, while the minimum was recorded on \textbf{17 January 1982}, with a water volume of \textbf{1{,}099{,}467~m\textsuperscript{3}}. It is worth noting that, according to LRA records, the reported water volume for that period is \textbf{66{,}380{,}000~m\textsuperscript{3}}, which is more than 60 times higher than our estimated value. Upon revisiting the satellite image for that specific date shown in Figure~\ref{fig:min}, it is evident that the water elevation was approximately 820 m, and the lake was nearly empty; therefore, our estimate appears to be far more realistic than the LRA measurement.

Quantitative analysis was conducted by evaluating several error metrics to assess different aspects of the SVR model’s overall performance relative to the LRBMS volume measurements. The employed metrics include the Mean Absolute Error ($MAE$), Root Mean Standard Deviation Ratio ($RSR$), Mean Absolute Percentage Error ($MAPE$), Root Mean Square Error ($RMSE$), Coefficient of Determination ($R^2$), and Percent Bias ($PBIAS$), selected for their relevance to near–real-time water volume monitoring applications. The $MAE$ and $RMSE$ quantify prediction errors, $MAPE$ represents relative accuracy, while $RSR$ contextualizes the error spread relative to natural variability. The $R^2$ values indicate the model’s ability to explain observed fluctuations, and $PBIAS$ assesses tendencies in systematic prediction bias.

As presented in Table~\ref{tab:landsat_results}, all metrics meet or exceed established benchmarks for SVR-based volume estimation using Landsat imagery alone. In 2023, the model achieved an $MAE$ of 5.10 and an $RMSE$ of 5.70, while in 2024 the $MAE$ increased slightly to 7.50 and the $RMSE$ to 10.50. These values remain well within acceptable limits, reflecting negligible absolute deviations relative to reservoir storage capacity. Both years achieved $MAPE$ values below 6\% (4.72\% in 2023 and 5.69\% in 2024), classifying the forecasts as “highly accurate”~\cite{6}. The $RSR$ values remained well below the 0.7 threshold (0.216 and 0.203 in 2023 and 2024, respectively), indicating statistically robust residual distributions~\cite{7}. Coefficients of determination ($R^2$) of 0.989 and 0.985 greatly exceed the 0.7 benchmark for satisfactory environmental models, explaining over 98\% of the variance in reservoir volume~\cite{8}. Finally, $PBIAS$ values of –4.57\% and –5.14\% fall comfortably within the ±25\% range, demonstrating the absence of significant systematic error~\cite{6}.

\begin{figure*}[t]
    \centering
    \includegraphics[scale=0.15]{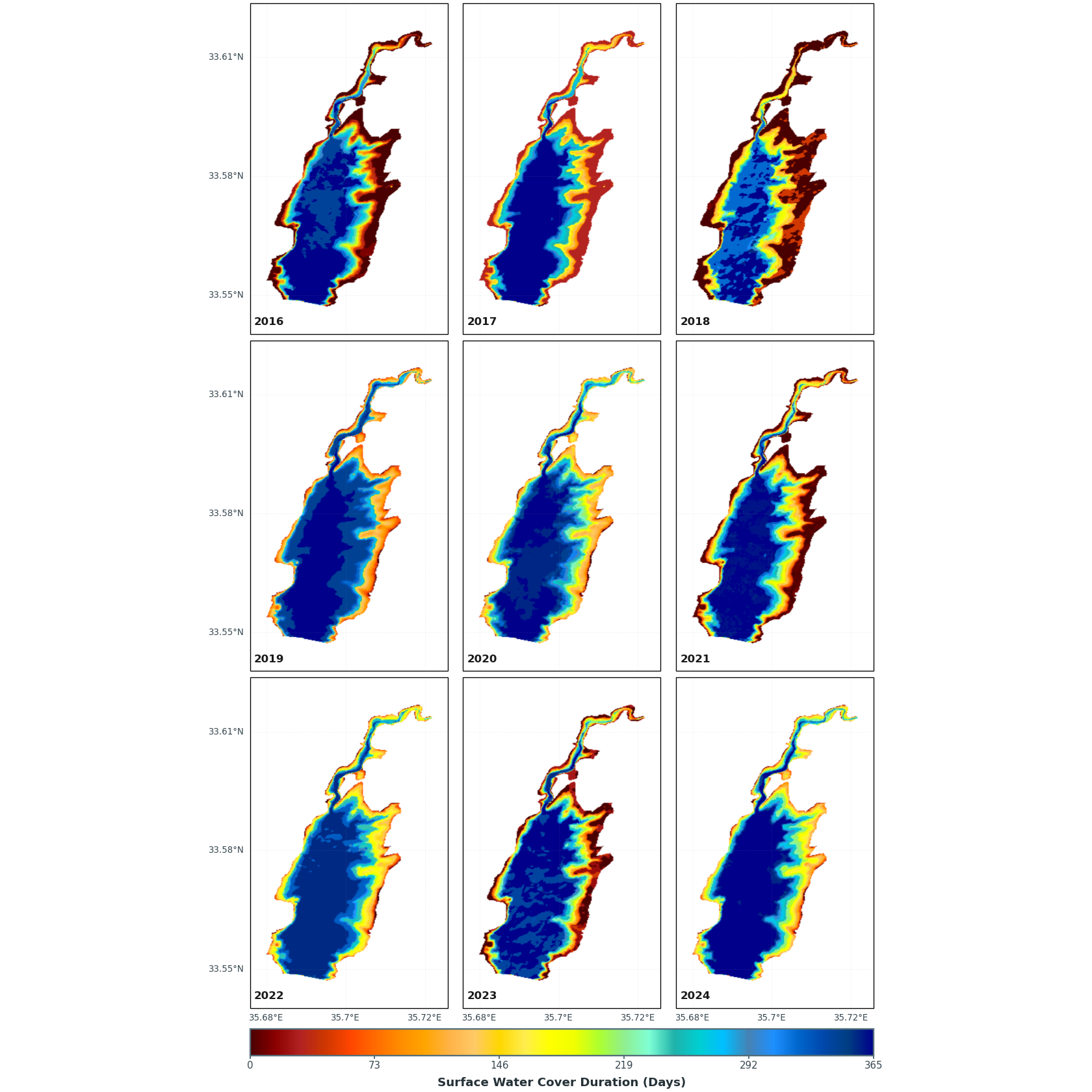}
    \caption{Water persistence map of the Qaraaoun Reservoir (2016–2024), derived from Sentinel-2 imagery. Blue areas indicate permanent water coverage (>300 days), while yellow to red areas represent ephemeral or seasonal water presence (<150 days). 
    The map reveals a stable reservoir core alongside variable shoreline zones, consistent with documented hydrological variability.}
    \label{fig:water persistence map}
\end{figure*}

\begin{table}[H]
\centering
\begin{tabular}{|l|c|c|}
\hline
\textbf{Metric} & \textbf{2023} & \textbf{2024} \\ \hline
MAE ($10^6\,\mathrm{m}^3$) & 5.1 & 7.5 \\ \hline
RMSE ($10^6\,\mathrm{m}^3$) & 5.7 & 10.5 \\ \hline
MAPE (\%) & 4.72 & 5.69 \\ \hline
RSR & 0.216 & 0.203 \\ \hline
R² & 0.989 & 0.985 \\ \hline
PBIAS (\%) & -4.57 & -5.14 \\ \hline
\end{tabular}
\caption{Volume estimation accuracy metrics for \textbf{Landsat} Imagery in years 2023 and 2024}
\label{tab:landsat_results}
\end{table}

For SVR estimations based on Sentinel-2 imagery, Table~\ref{tab:sentinel_results} likewise demonstrates consistent model performance in both 2023 and 2024, maintaining alignment with scholarly benchmarks. In 2023, the model achieved an $MAE$ of 4.53 and an $RMSE$ of 5.22, which increased in 2024 to $MAE$ = 7.08 and $RMSE$ = 9.85—values that remain negligible in practical terms. Both years produced $MAPE$ values below 6\% (4.28\% in 2023 and 5.22\% in 2024), classifying them again as “highly accurate” forecasts~\cite{7}. The $RSR$ values remained well under 0.7 (0.194 and 0.200 in 2023 and 2024, respectively), indicating statistically robust residual distributions~\cite{6}. The coefficients of determination ($R^2$) of 0.988 and 0.983 far exceed the 0.7 benchmark, explaining more than 98\% of volume variance~\cite{8}. Finally, $PBIAS$ values of –4.11\% and –4.74\% fall well within the ±25\% “satisfactory” range, confirming the absence of systematic bias~\cite{6}.

\begin{table}[H]
\centering
\begin{tabular}{|l|c|c|}
\hline
\textbf{Metric} & \textbf{2023} & \textbf{2024} \\ \hline
MAE (10\textsuperscript{6} m³) & 4.53 & 7.08 \\ \hline
RMSE (10\textsuperscript{6} m³) & 5.22 & 9.85 \\ \hline
MAPE (\%) & 4.28 & 5.22 \\ \hline
RSR & 0.194 & 0.2 \\ \hline
R² & 0.988 & 0.983 \\ \hline
PBIAS (\%) & -4.11 & -4.74 \\ \hline
\end{tabular}
\caption{Volume estimation accuracy metrics for \textbf{Sentinel-2} Imagery in years 2023 and 2024}
\label{tab:sentinel_results}
\end{table}


The qualitative and quantitative results demonstrate that the proposed solution pipeline reliably segments and predicts Qaraaoun Reservoir volumes across different sensors and years. Segmentation aligns with ground truth along more than 95\% of the shoreline, while volume estimates exhibit relative errors below 6\%, $RSR \approx 0.2$, $R^2 > 0.98$, and negligible bias—exceeding established hydrologic and forecasting benchmarks. These findings confirm the tool’s suitability for cost-effective, sensor-free operational monitoring of reservoir storage.

Furthermore, a multi-year water persistence analysis (2016--2024) was conducted to evaluate the spatial stability of surface water coverage in the reservoir, as depicted in Fig.~\ref{fig:water persistence map}. The persistence map reveals a stable central core that remains water-covered for more than 300 days annually, while shoreline regions exhibit high variability, with coverage durations often below 150 days. These fluctuating margins correspond to seasonal inflows and interannual droughts, reinforcing the temporal variability observed in the long-term volume record.




\section{Conclusion}
This study proposes an innovative approach for real-time monitoring of the Qaraaoun Reservoir’s water volume through the utilization of open-source satellite imagery, supported by a robust and well-documented pipeline for water extent detection and advanced machine learning algorithms. Through comprehensive analysis, highly accurate segmentation and reliable volume estimations were achieved. The results demonstrate that the proposed methodology enables a cost-effective, sensor-free solution characterized by minimal error and strong resilience to environmental challenges. 

Despite its promising outcomes, the approach presents certain limitations inherent to the reliance on satellite data, such as atmospheric interference, which directly affects segmentation accuracy. Furthermore, the selected water index must be carefully adapted to local geographic conditions, seasonal variations, and the specific hydrological mechanisms governing inflows into the reservoir. Nevertheless, by combining multiple indices and enhancing the inference model, this study has paved the way toward durable and scalable monitoring systems.

The Qaraaoun Reservoir represents a vital resource for Lebanon’s agriculture (supporting more than 40{,}000~ha) and domestic water supply, both of which stand to benefit significantly from this proposed tool—particularly given the country’s limited monitoring infrastructure and frequent sensor malfunctions. This research underscores the potential of remote sensing and machine learning to address local challenges and to advance sustainable water management practices in developing regions. Moving forward, further optimization of the algorithm and inference model, together with the integration of more extensive real-time datasets, could enhance the system’s long-term sustainability and operational resilience. Additionally, incorporating water quality indices derived from Landsat and Sentinel imagery would represent a valuable extension to the proposed dashboard, enriching its analytical and environmental monitoring capabilities.

\section*{Acknowledgments}
This work was partially supported by the first Call for Proposals for Researchers - SEALACOM project.

\bibliography{references}

\end{document}